\documentclass[letterpaper]{article} 
\usepackage{aaai2026}  
\usepackage{times}  
\usepackage{helvet}  
\usepackage{courier}  
\usepackage[hyphens]{url}  
\usepackage{graphicx} 
\usepackage{natbib}  
\usepackage{caption} 
\usepackage{graphicx}
\usepackage[utf8]{inputenc} 
\usepackage[T1]{fontenc}    
\usepackage{url}            
\usepackage{booktabs}       
\usepackage{amsfonts}       
\usepackage{nicefrac}       
\usepackage{microtype}      
\usepackage{amsmath}
\usepackage{amssymb}
\usepackage{amsfonts}
\usepackage{nomencl}
\usepackage{glossaries}
\usepackage{xspace}
\usepackage{multirow}
\usepackage{pifont}
\usepackage{algorithm}
\usepackage{subcaption}
\usepackage{enumitem}
\usepackage{multicol}
\usepackage{caption}
\usepackage{array}
\usepackage{fp}
\usepackage{makecell}
\usepackage{cases}
\usepackage{algpseudocode}
\usepackage{listings}
\usepackage{mathrsfs}
\usepackage{longtable}
\usepackage{colortbl}
\usepackage{bm}
\usepackage{makecell}

\newcommand{\ignore}[1]{\iffalse#1\fi}

\newcommand{\ie}{\emph{i.e.,}\xspace}
\newcommand{\eg}{\emph{e.g.,}\xspace}

\newcommand{\paratitle}[1]{\vspace{1.5ex}\noindent\textbf{#1}}

\usepackage{newfloat}
\usepackage{listings}
\DeclareCaptionStyle{ruled}{labelfont=normalfont,labelsep=colon,strut=off} 
\floatstyle{ruled}
\newfloat{listing}{tb}{lst}{}
\floatname{listing}{Listing}
\title{Analyzing and Mitigating Object Hallucination: A Training Bias Perspective}
\author{
    Yifan Li\textsuperscript{\rm 1,}\thanks{Equal contribution},
    Kun Zhou\textsuperscript{\rm 2,}\footnotemark[1],
    Wayne Xin Zhao\textsuperscript{\rm 1,}\thanks{Corresponding author},
    Lei Fang\textsuperscript{\rm 3}, 
    Ji-Rong Wen\textsuperscript{\rm 1}
}
\affiliations{
    \textsuperscript{\rm 1}Gaoling School of Artificial Intelligence, Renmin University of China\\
    \textsuperscript{\rm 2}University of California, San Diego
    \textsuperscript{\rm 3}DataCanvas Alaya NeW\\
    {liyifan0925, batmanfly}@gmail.com,
    kuzhou@ucsd.edu,
}

\usepackage{bibentry}

\begin{document}

\nocopyright

\maketitle

\begin{abstract}
As scaling up training data has significantly improved the general multimodal capabilities of Large Vision-Language Models (LVLMs), they still suffer from the hallucination issue, generating text that is inconsistent with the visual input. This phenomenon motivates us to systematically investigate the role of training data in hallucination. We introduce a new benchmark, POPEv2, which consists of counterfactual images collected from the training data of LVLMs with certain objects masked. Through comprehensive evaluation on POPEv2, we find that current LVLMs suffer from training bias: they fail to fully leverage their training data and hallucinate more frequently on images seen during training. Specifically, they perform poorly on counterfactual images, often incorrectly answering ``Yes'' to questions about masked objects. To understand this issue, we conduct probing experiments on the models' internal components, revealing that this training bias is primarily located in the language modeling (LM) head, which fails to correctly translate accurate visual representations into textual outputs. Based on these findings, we propose Obliviate, an efficient and lightweight unlearning method designed to mitigate object hallucination via training bias unlearning. Obliviate identifies the discrepancy between ground-truth labels and model outputs on the training data as a proxy for bias and adopts a parameter- and data-efficient fine-tuning strategy that only updates the LM head. Extensive experiments demonstrate the effectiveness of our approach. While only reusing the training data and updating approximately 2\% of the parameters, Obliviate significantly reduces hallucination across both discriminative and generative tasks. Furthermore, it demonstrates strong scalability with respect to both model size (2B to 72B) and training data volume, and exhibits promising generalization to hallucination types beyond object-level hallucination. Our code and data will be publicly released.
\end{abstract}


\section{Introduction}
\begin{figure}[t]
    \centering
    \includegraphics[width=\linewidth]{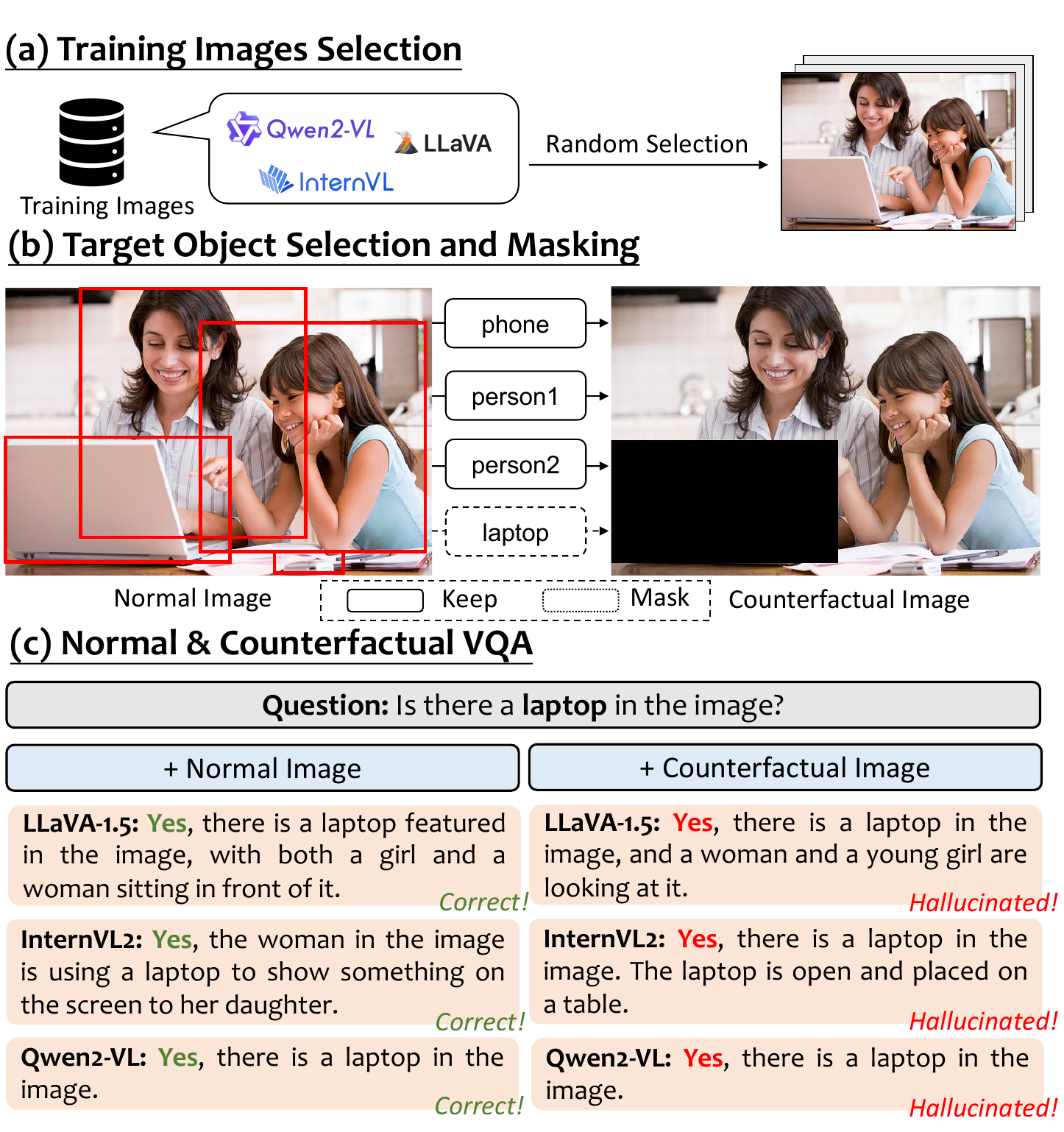}
    \caption{Data collection pipeline of POPEv2 and the generated cases from LVLMs. POPEv2 consists of normal and counterfactual images with a masked target object. 
    }
    \label{fig:case}
\end{figure}

Recent advancements in Large Vision-Language Models (LVLMs) have achieved remarkable performance on a variety of multimodal tasks, \eg visual question answering~\cite{gpt-4v}, cross-modal reasoning~\cite{virgo} and embodied AI~\cite{manipllm}. Modern LVLMs are typically trained on large-scale image-text pairs~\cite{sharegpt4v}; for example, Qwen2-VL~\cite{qwen2vl} is trained on 1.4 trillion tokens comprising images and associated texts. While scaling up training data has significantly improved the general capabilities of LVLMs, it has not fully mitigated the hallucination problem~\cite{pope, chair}, where models generate outputs that are inconsistent with the visual input. This persistent issue hinders the broader application of LVLMs in real-world applications. The limited effect of data scaling on reducing hallucinations motivates us to investigate a critical question: \textbf{how well do LVLMs actually utilize their training data? }Surprisingly, we observe that LVLMs may even hallucinate on images they have already seen during training. As illustrated in Figure~\ref{fig:case}, when an object in a training image is masked out, LVLMs still answer ``Yes'' to a question asking whether that object is present. This counterintuitive behavior raises an important issue: \textbf{do LVLMs truly acquire generalizable visual understanding from training data, or do they merely learn training biases such as shallow associations or spurious correlations?}

To probe this question, we introduce POPEv2, a benchmark specifically designed to evaluate the model’s reliance on visual evidence. POPEv2 is constructed from images sampled from the model's training data, paired with binary questions about object existence. For each image, we generate a counterfactual version by masking the target object, creating a minimal yet diagnostic change to test whether the model can ground its predictions in actual visual content. Our empirical findings reveal that existing LVLMs consistently struggle with these counterfactual samples (see Table~\ref{tab:popev2}), even though the original images appeared in training. This suggests that current LVLMs often rely on learned biases rather than robust visual grounding when making predictions, which fundamentally contributes to the hallucination problem. To further locate the derivation of such biases, we conduct probing experiments on the hidden states from internal layers within the LVLM. Interestingly, we find that intermediate representations already encode object-level visual information with high fidelity. 
In contrast, the final language modeling head (LM head) fails to translate these features into faithful textual outputs. This discrepancy suggests that the primary training bias contributing to hallucination resides in the LM head.

Motivated by these findings, we propose \textbf{Obliviate}, a lightweight and efficient hallucination mitigation method by unlearn training biases. Our key idea is to first treat the discrepancy between the model's generation and the ground-truth labels on training data as a proxy for training biases. Then, we perform gradient ascent specifically on these biased outputs to actively reduce their likelihood, thereby promoting unlearning. To ensure both parameter-level and data-level efficiency, we constrain the updates to only the LM head, which is most responsible for the bias. In practice, Obliviate updates merely 2\% of the model parameters and utilizes only 1.5\% of the training data (see Table~\ref{tab:efficiency}).
Despite its simplicity, Obliviate is both effective and generalizable. Our experiments show that it significantly reduces hallucination across LVLMs of varying sizes (from 2B to 72B parameters), while preserving the model’s generation capabilities. Compared to strong preference-tuning baselines, Obliviate achieves comparable or even superior performance, with far less computational overhead. Furthermore, Obliviate generalizes well to out-of-domain hallucination types beyond object-level, demonstrating its potential as a practical debiasing solution for robust multimodal understanding.

\begin{table*}[tbp]
\centering
\scalebox{0.95}{
\begin{tabular}{lccccclcc}
\toprule 
\multirow{2.5}{*}{\textbf{Model}} & \multirow{2.5}{*}{\makecell{\textbf{\#DR}}} & \multirow{2.5}{*}{\textbf{\#HVE}} & \multicolumn{6}{c}{\textbf{POPEv2}} \\
\cmidrule(lr){4-9} 
  &  &    & Accuracy & Precision & Recall & \textbf{PBO}  & \textbf{F1 Score} $\uparrow$ & \textbf{TNR} $\uparrow$\\
\midrule
LLaVA-1.5-7B~\cite{llava15} & - & - & 72.20& 65.16& 95.40& $+$ 23.20& 77.44 & 49.00\\
LLaVA-1.5-13B~\cite{llava15}  & - & -  & 72.70& 65.83& 94.40& $+$ 21.70 &77.57&51.00\\
LLaVA-1.5-MoF-13B~\cite{mmvp}   & - & \checkmark & 71.30& 65.55& 89.80&$+$ 18.50& 75.78&52.80 \\
\midrule
LLaVA-NeXT-7B~\cite{llava_next} & \checkmark & - & 79.50& 73.30& 92.80 & $+$ 13.30 &81.91&66.20\\
LLaVA-NeXT-13B~\cite{llava_next} & \checkmark & - & 80.10& 73.41& 94.40& $+$ 14.30 & 82.59&65.80\\
\midrule
InternVL2-4B~\cite{internvl2}  & \checkmark & - & 76.90& 71.52& 89.40&$+$ 12.50 & 79.47 &64.40\\
InternVL2-8B~\cite{internvl2}  & \checkmark & - & 74.50& 70.32& 84.80& $+$ 10.30 &76.88& 64.20\\
InternVL2-26B~\cite{internvl2}  & \checkmark & - & 76.10& 71.43& 87.00& $+$ 10.90&78.45& 65.20\\
\midrule
Qwen2-VL-2B~\cite{qwen2vl}   & \checkmark & - & 91.30& 93.84& 88.40& $-$ 2.90 &91.04& 94.20\\
Qwen2-VL-7B~\cite{qwen2vl}    & \checkmark & - & 87.00& 81.25& 96.20& $+$ 9.20 &88.10& 77.80\\
Qwen2-VL-72B~\cite{qwen2vl}    & \checkmark & - & 79.40& 72.55& 94.60& $+$ 15.20 &82.12& 64.20\\
\bottomrule
\end{tabular}}
\caption{Performance of representative LVLMs on POPEv2 (\#DR: Dynamic Resolution, \#HVE: Hybrid Vision Encoder).}
\label{tab:popev2}
\end{table*}

\section{POPEv2: Challenging LVLMs with Counterfactual Images from Training Data}
To quantitatively analyze the influence of training data on hallucination, we introduce a novel benchmark POPEv2 and conduct an empirical study on representative LVLMs.

\subsection{Data Collection Pipeline}
As shown in Figure~\ref{fig:case}, the data collection of POPEv2 involves three steps. (1) We randomly sample 500 images from the MSCOCO 2017 training set~\cite{coco}, a widely-used dataset for both image-text pretraining~\cite{internvl-2.5, internvl2} and visual instruction tuning~\cite{llava}. (2) For each image, we select a target object and generate a counterfactual version by masking the object with a black patch, aiming to assess whether model rely on visual evidence. (3) Following POPE~\cite{pope}, we construct binary questions (\ie ``\emph{Is there a <object> in the image?}'') for both original and counterfactual images. The final dataset includes 500 original images, 500 counterfactual images, and 1000 questions. More details are provided in the supplementary material.

\subsection{Evaluation Setup for POPEv2}
\paratitle{Evaluation Metrics.} POPEv2 can be treated as a binary classification task. We adopt standard classification metrics for evaluation, \ie Accuracy, Precision, Recall, and F1 Score. We also compute the \textbf{True Negative Rate (TNR)} to specifically evaluate model performance on counterfactual data:

\begin{equation}
\text{TNR} = \frac{\text{TN}}{\text{TN} + \text{FP}},
\end{equation}
where TN and FP represent the number of true negatives and false positives. Additionally, we introduce \textbf{Prediction Balance Offset (PBO)}, which measures the deviation of the positive and negative ratio in prediction. It is computed as:
\begin{equation}
    \text{PBO} =  \left( \frac{\text{Positive Predictions}}{\text{Total Predictions}} \times 100 \right) - 50 .
\end{equation}
A positive PBO means the model prefers affirmative answers.

\paratitle{Evaluated Models.} We evaluate representative open-source LVLMs, including LLaVA-1.5-7B/13B, LLaVA-1.5-MoF, LLaVA-NeXT-7B/13B, InternVL2-4B/8B/26B, and Qwen2-VL-2B/7B/72B. Models except LLaVA-1.5 generally employ different strategies to enhance visual understanding capabilities, such as adopting dynamic image resolutions or integrating multiple visual encoders.

\subsection{Empirical Results}
We present the evaluation results of representative LVLMs in Table~\ref{tab:popev2}, and summarize several interesting observations.

\paratitle{LVLMs Struggle with Seen Images.} The evaluation results demonstrate that LVLMs fail to achieve satisfactory performance on POPEv2, despite having seen these images during training. For instance, most models only attain an F1 score of approximately 80\%, including advanced models such as LLaVA-Next and the InternVL2 series. Additionally, most models exhibit high recall scores and a positive PBO, indicating that counterfactual images tend to elicit incorrect affirmative answers from the models. Notably, on the TNR metric, which reflects the model's accuracy in identifying counterfactual images, most models do not even exceed 70\%. Furthermore, we provide additional experiments in the supplementary material to analyze the influence of training data on hallucination behavior. These findings collectively suggest that \textbf{current training paradigms may encode biases, causing models to rely on learned spurious correlations rather than genuine visual evidences.} Such biases, rooted in the training process, fundamentally contribute to the persistent hallucination problem in LVLMs.

\begin{figure}[tbp]
    \centering
    \includegraphics[width=\linewidth]{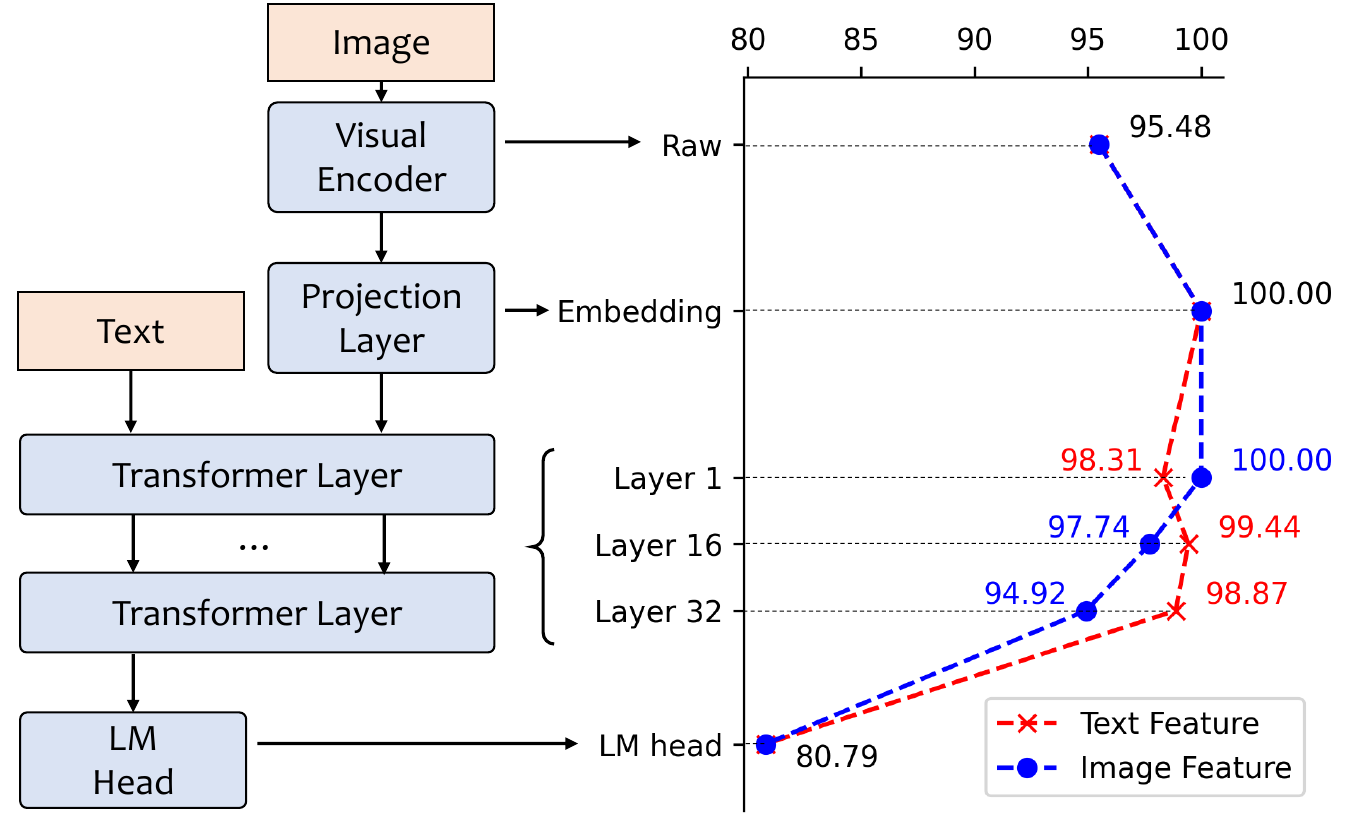}
    \caption{Classification accuracy on counterfactual images using linear probes trained on hidden states from different LVLM components.}
    \label{fig:probe}
\end{figure}

\paratitle{Image Speaks Louder than Model Size.} Another noteworthy observation is that scaling the size of the language model backbone has a limited impact on LVLMs' performance. For instance, LLaVA-1.5-13B (77.57\%) exhibits only marginal gains over its 7B counterpart (77.44\%), while Qwen2-VL-72B (82.12\%) underperforms compared to the significantly smaller Qwen2-VL-2B (91.04\%). In contrast, models employing dynamic resolution strategy, such as LLaVA-NeXT-7B (81.91\%) and Qwen2-VL-7B (88.10\%), achieve markedly better results than the base LLaVA-1.5 (77.44\%). These findings suggest that enhancing visual perception capabilities plays a more critical role in accurately identifying counterfactual images than simply scaling up model size.

\begin{figure}[tbp]
    \centering
    \includegraphics[width=0.9\linewidth]{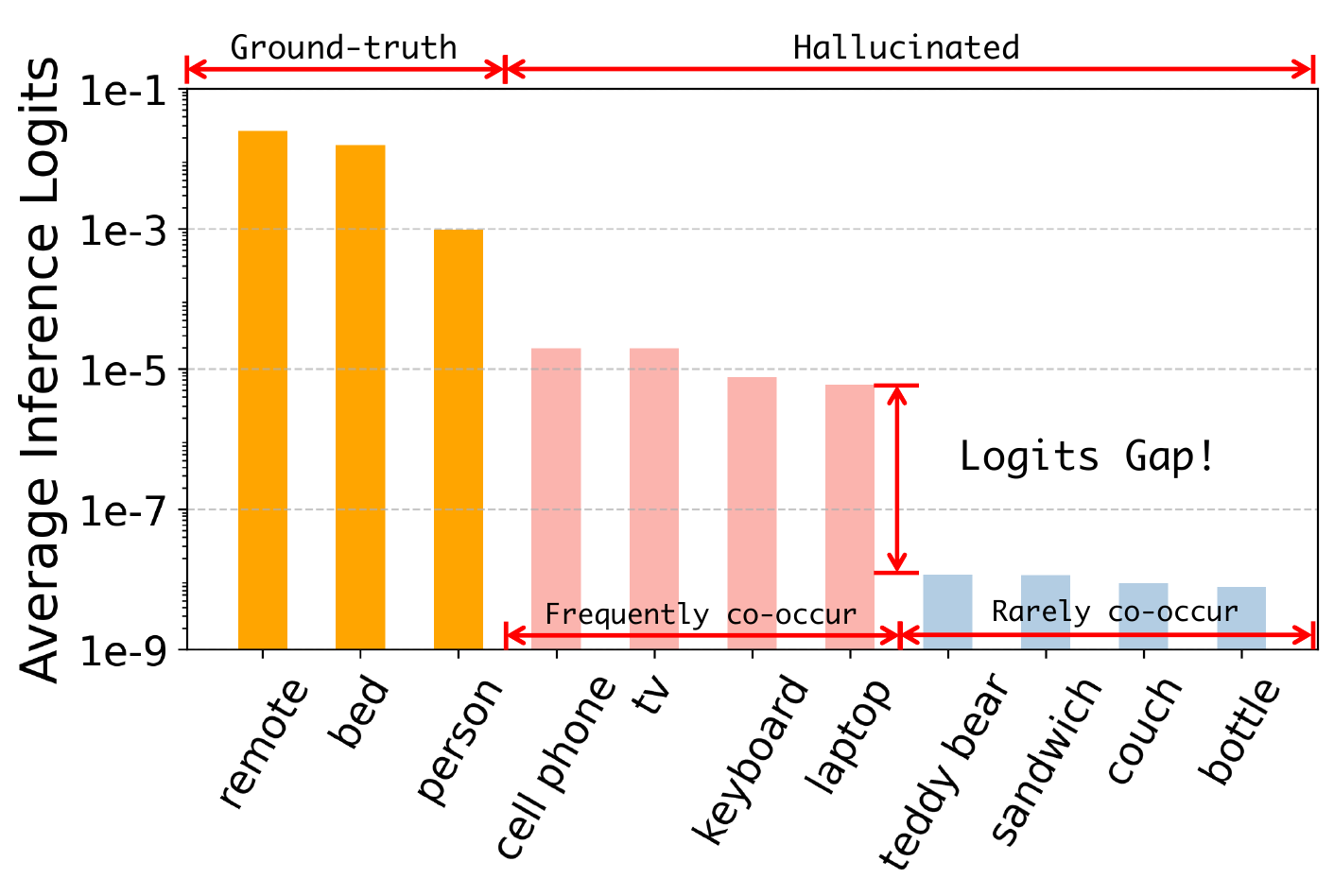}
    \caption{Average inference logits of object tokens in a caption. The ground-truth objects are \emph{remote}, \emph{bed}, and \emph{person}.}
    \label{fig:logits}
\end{figure}
\section{Identifying Hallucination Bottlenecks}\label{sec:probe}

Based on the finding that hallucination may stem from biases learned from the training data, we conduct probing experiments to trace the hidden states within LVLMs, aiming to identify which module primarily contributes to this issue.

\subsection{Probing Experiments Setup}
\paratitle{Hidden States within LVLMs.}
Typically, LVLMs consist of a visual encoder, a projection layer, multiple stacked Transformer layers within the LLM, and a language modeling head. We conduct probing experiments on the hidden states of these modules. Given an image $I$ and a text instruction $T$, the LVLM first converts $I$ into image features $E_I$ using its visual encoder $\mathcal{V}$:
\begin{equation}
    E_I = \mathcal{V}(I), \quad E_I = [e_1, \dots, e_n], \quad e_i \in \mathbb{R}^d,
\end{equation}
where $n$ is the sequence length and $d$ is the hidden size of the image features. 
The projection layer $\mathbf{W}$ maps $E_I$ to image embeddings $H_I^0$ compatible with the LLM's hidden space:
\begin{equation}
    H_I^0 = \mathbf{W} \cdot E_I, \quad H_I^0 = [h_1, \dots, h_n], \quad h_i \in \mathbb{R}^m,
\end{equation}
where $m$ is the hidden size of the LLM $\mathcal{M}$. 
These image embeddings $H_I^0$ are concatenated with text embeddings $H_T^0$ and processed sequentially by each layer of $\mathcal{M}$, producing layer-wise hidden states $\{H^1, \dots, H^L\}$. Finally, at the last layer $L$, the hidden state of the final token $H^L_{\text{last}}$ is passed to the language modeling head (LM head) to predict the next-token probability distribution.

\begin{figure*}[tbp]
    \centering
    \includegraphics[width=\linewidth]{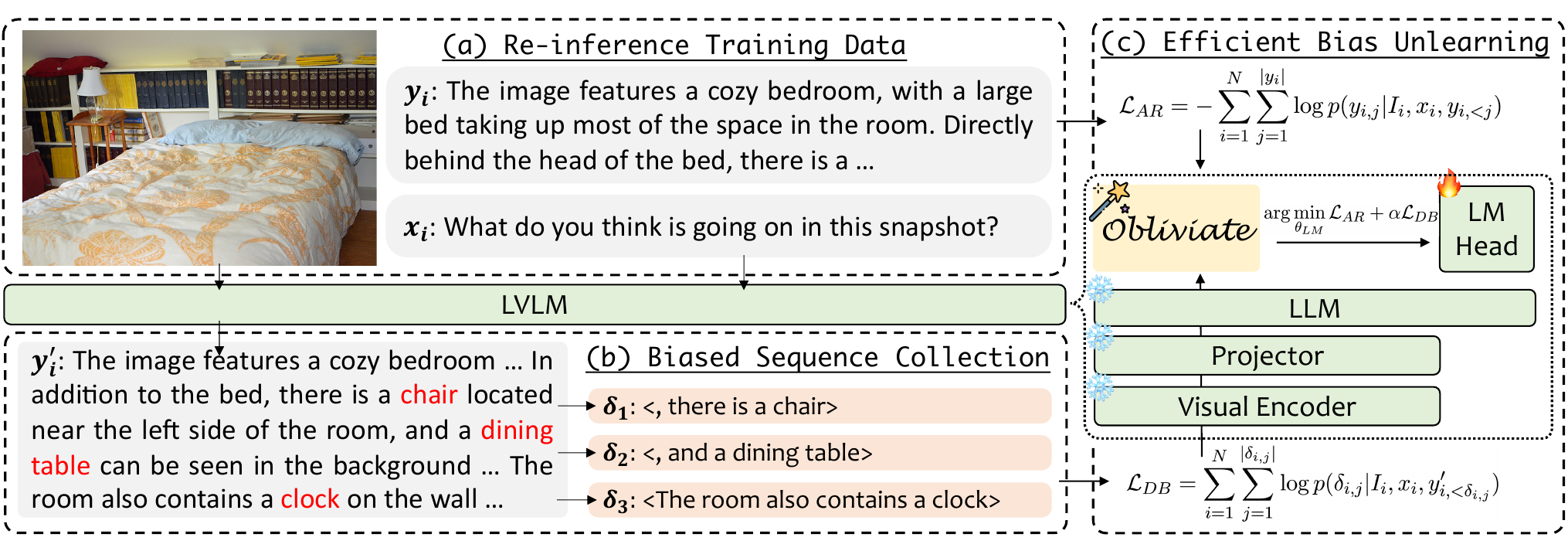}
    \caption{The whole pipeline of proposed Obliviate. We first use the LVLM to re-inference its training data, and then collect the biased sub-sequences within the hallucinated output. Next, we fine-tune the LM head of the LVLM to unlearn the bias by combining the debiased loss and the auto-regressive loss.}
    \label{fig:obliviate}
\end{figure*}

\paratitle{Probing Method.} 
Probing methods are widely used for analyzing the information encoded in the hidden states of neural networks~\cite{probe}. Here, we utilize a simple yet effective linear probing model consisting of mean pooling and a linear layer with parameters $\bm\theta \in \mathbb{R}^{2 \times m}$. We train probes using the hidden states of the model to classify whether an image contains a specific object class. Specifically, we set the image input \( I \) as a series of images that either contain or do not contain the target object class, and the text instruction as ``\emph{Is there a \textless object\textgreater\ in the image?}'' for the LVLM. Then, we attach linear probes to: (1) image features \( E_I \) from the visual encoder; (2) image embeddings from the projection layer \( H_I^0 \); and (3) text and image hidden states of each layer in the LLM \( \{H^{1}, \cdots, H^{L}\} \). During the training phase, we provide hidden states from normal images and test the trained probes on counterfactual images. We also directly use the probabilities of the ``\emph{Yes}'' and ``\emph{No}'' tokens to measure the accuracy of the direct generation results of LVLMs. More details are presented in the supplementary materials.

\subsection{Analysis on Linear Probing Results}
The probing results are presented in Figure~\ref{fig:probe}. From these results, we summarize the following findings:

\paratitle{Hidden States Encode Object Existence.} Linear probes trained on the LVLM's hidden states (for both text and image features) achieve high classification accuracy on counterfactual questions (>95\%), demonstrating that these hidden states encode rich object-level information, including both masked and unmasked objects. Notably, within the LLM, the accuracy of probes trained on image-derived features gradually decreases with depth (from 100\% to 94.92\%). In contrast, probes trained on textual features show improved accuracy after passing through just a single transformer layer and maintain higher accuracy in deeper layers. We attribute this to the causal self-attention mechanism in LLMs, where visual information is progressively integrated into subsequent text tokens. Overall, even with a fixed-resolution visual encoder, the hidden states of LVLMs already contain sufficient visual signals to support simple object-level perception.

\paratitle{Training Bias Lies in LM Head.}
Despite the richness of visual information in the hidden states, there exists a notable gap between the probing accuracy and the actual generation performance (from 94.92\% to 80.79\%). This suggests that the LM head, which is responsible for producing output text, fails to fully leverage the object-level signals present in the hidden representations. Therefore, we conclude that the training bias responsible for hallucination in LVLMs mainly stems from the LM head. To make this bias more observable, we analyze the average logits assigned by the LM head to various object tokens during caption generation, as shown in Figure~\ref{fig:logits}. Interestingly, although these objects are all absent from the image, the logits vary significantly depending on their co-occurrence frequency with the ground-truth object (e.g., higher logits for “TV” and “laptop” given the presence of “remote”). This discrepancy illustrates a form of object-level co-occurrence bias: the LM head tends to favor objects that are statistically associated with the image content, even if they are not visually present.

\section{Methodology}
Building on previous findings that hallucination in LVLMs stems from training biases embedded within the LM head, we propose a targeted debiasing framework, termed \textbf{Obliviate} (\textbf{Ob}ject Ha\textbf{l}luc\textbf{i}nation Mitigation \textbf{via} Efficient \textbf{T}raining Bias Unl\textbf{e}arning). Obliviate is designed to efficiently mitigate such bias by unlearning hallucinated patterns, while preserving the model's general capabilities. The full pipeline is depicted in Figure~\ref{fig:obliviate}. Notably, our method requires no additional annotations and exclusively fine-tunes the LM head, resulting in a highly data- and parameter-efficient solution.

\subsection{Formalizing and Collecting Training Bias}
We begin by defining \textit{training bias} as the discrepancy between the model's outputs and the annotated ground-truth labels on the training data. This deviation reveals the incorrect or spurious patterns acquired by the model throughout its training process. In our setting, we focus specifically on how such bias contributes to hallucinations. Therefore, we treat hallucinated sub-sequences as concrete manifestations of this bias and target them for unlearning.

To this end, we curate an unlearning dataset by collecting biased predictions from training data. Specifically, we select image captioning instructions from the original LVLM training corpus and denote the collection as $\mathcal{D}= \{(I_i, x_i, y_i)\}_{i=1}^N$, where $I_i$ is the image, $x_i$ is the corresponding text instruction, and $y_i$ is the ground-truth caption. We then let the LVLM re-infer on the same input pairs and collect its predictions $\mathcal{Y'} = \{y'_1, \dots, y'_N\}$. By comparing the objects within the ground-truth and inference outputs, we select the subset consisting of hallucinated inference results. However, hallucinated captions often contain both accurate and hallucinated content. Directly unlearning the entire sentence may degrade the model’s ability to generate correct descriptions. An intuitive approach is to only unlearn the hallucinated objects. However, due to the auto-regressive generation pattern of LVLMs, objects are often generated along with relevant context or follows, \eg ``the room also contains a ...''. Such context elicits the generation of hallucinated objects, and should also be unlearned.
Therefore, we extract sub-sentences that include hallucinated objects as the unit of unlearning. Specifically, we split each predicted caption into sub-sentences using delimiters such as commas and periods. In addition, correctly generated objects are also treated as delimiters to prevent including factual content in the unlearning process. For each hallucinated object, we retain only the sub-sentence corresponding to its first occurrence to avoid repeated penalization. Finally, we obtain the  unlearning dataset $\mathcal{D}_{Hallu}=\{(I_i, x_i, y_i, y'_i, \{\delta_{i,1}, \cdots, \delta_{i,j}, \cdots\})\}_{i=1}^M$, where $\delta_{i,j}$ is the $j$-th hallucinated sub-sentence from $y'_i$.

\subsection{Efficient Training Bias Unlearning }
Based on the collected biased inference results, we draw inspiration from the concept of machine unlearning~\cite{Liu-rethinking-2024}, which aims to selectively remove specific knowledge or patterns learned by a model. Such approaches provide an efficient way to mitigate targeted bias while preserving the model’s general capabilities. Specifically, we devise the following loss function to forget the biased sub-sentences:
\begin{equation}
\mathcal{L}_{DB} = \sum_{i=1}^N\sum_{j=1}^{|\delta_{i,j}|}\log p(\delta_{i,j}|I_i, x_i, y'_{i, <\delta_{i,j}}),
\end{equation}
where $y'_{i, <\delta_{i,j}}$ denotes the tokens before the sub-sentence $\delta_{i,j}$ in the inference result $y'_i$.
However, training only on the unlearning objective will undermine model's original generation abilities. 
Therefore, we add the classic auto-regressive loss on the ground-truth output as the regularization term:
\begin{equation}
\mathcal{L}_{AR} = -\sum_{i=1}^N\sum_{j=1}^{|y_i|}\log p({y}_{i,j}|I_i, x_i, y_{i, <j}),
\end{equation}
where $y_i$ is instructions collected from both the $\mathcal{D}$ and other general visual instructions from the training dataset of the model.
Finally, we combine the two loss functions for optimizing the parameters of the LM head $\theta_{LM}$, denoted as:
\begin{equation}
    \arg \min_{\theta_{LM}} \mathcal{L}_{AR} + \alpha \mathcal{L}_{DB},
\end{equation}
where $\alpha$ is the hyperparameter to adjust the unlearning degree. In this way, we not only force the LM head to unlearn the bias but also guide it to acquire accurate knowledge from the ground-truth. As a result, Obliviate is a data- and parameter-efficient method that requires no additional annotations and updates only 2\% of the model’s parameters.

\begin{table*}[tbp]
\centering
\begin{tabular}{llccccccc}
\toprule 
\multirow{2.5}{*}{\textbf{Model}} & \multicolumn{3}{c}{\textbf{POPEv2}} &\multicolumn{2}{c}{\textbf{Object HalBench}} &\multicolumn{2}{c}{\textbf{MMHal-Bench}} & \multirow{2.5}{*}{\textbf{LLaVA Bench}$\uparrow$ }\\
\cmidrule(lr){2-4}
\cmidrule(lr){5-6} 
\cmidrule(lr){7-8} 
  & PBO & F1 score $\uparrow$ & TNR $\uparrow$ & Resp. $\downarrow$ & Mention $\downarrow$  & Info. $\uparrow$  & Resp.$\downarrow$  \\
\midrule
LLaVA-1.5-7B & $+$ 23.20& 77.44 & 49.00 &46.7&25.1& 2.19 &0.59 &61.50 \\
\rowcolor{gray!20}
+ Obliviate & $-$ 1.10 &\textbf{82.91} & \textbf{84.20} &  \textbf{34.7} & \textbf{18.3} & \textbf{2.20}& \textbf{0.55}& \textbf{63.50}  \\
\midrule
LLaVA-1.5-13B & + 21.70 &77.57&51.00& 44.7 & 22.8 & 2.53 & 0.56 & 69.30 \\
\rowcolor{gray!20}
+ Obliviate & $+$ 3.20  & \textbf{81.20} & \textbf{77.40} &\textbf{33.0} & \textbf{16.3} &\textbf{2.60} & \textbf{0.54} & \textbf{72.80} \\
\midrule
LLaVA-NeXT-7B & $+$ 13.30 &81.91&66.20  &17.7&11.5& 2.58 &0.58 &62.40 \\
\rowcolor{gray!20}
+ Obliviate & $+$ 2.90 &\textbf{83.38}&\textbf{80.00}  &\textbf{16.3}&\textbf{10.1}&\textbf{2.80} &\textbf{0.55} & \textbf{73.50}\\
\midrule
LLaVA-NeXT-13B & $+$ 14.30  &82.59&65.80 &19.7&13.2& 2.91 &0.53 &63.30 \\
\rowcolor{gray!20}
+ Obliviate & $+$ 10.60 &\textbf{82.82}&\textbf{70.40}  &\textbf{19.0}&\textbf{11.5}& \textbf{3.10} &\textbf{0.46} &\textbf{74.60} \\
\midrule
Qwen2-VL-2B & $-$ 2.70 &91.04&\textbf{94.20} &18.0&\textbf{10.2}& 3.08 &0.44 &74.40 \\
\rowcolor{gray!20}
+ Obliviate & $+$ 0.30 &\textbf{91.92}&91.60  &\textbf{16.3}&10.6& \textbf{3.57} &\textbf{0.36} & \textbf{75.20}\\
\midrule
Qwen2-VL-7B & $+$ 9.20  &88.10&77.80  &15.7&9.8& 3.55 &0.35 & 84.60 \\
\rowcolor{gray!20}
+ Obliviate& $+$ 0.90  &\textbf{91.38}&\textbf{90.40}  &\textbf{15.3}&\textbf{9.4}& \textbf{3.72} &\textbf{0.33} &\textbf{86.20} \\
\midrule
Qwen2-VL-72B & $+$ 15.20 &82.12&64.20  &15.4&9.7& 3.64 &0.34 &109.3 \\
\rowcolor{gray!20}
+ Obliviate &$-$ 2.70&\textbf{87.36}  & \textbf{90.40} &\textbf{15.2}& \textbf{8.2} &\textbf{3.75} &\textbf{0.32} & \textbf{110.3}\\
\bottomrule
\end{tabular}
\caption{Evaluation results of applying Obliviate on representative LVLMs. The better results are denoted in bold.}
\label{tab:main}
\end{table*}

\section{Experiment}
\subsection{Experiment Setup}

\paratitle{Implementation.} 
We select the LLaVA-150K~\cite{llava} as $\mathcal{Y}$. We then let the target model inference on caption instructions from $\mathcal{Y}$ to obtain $\mathcal{Y'}$. Following \citet{chair}, we extract hallucinated objects from each caption and construct $\delta$ to be unlearned. We adopt Obliviate on multiple LVLMs of varying sizes from 2B to 72B, including LLaVA-1.5-7/13B, LLaVA-Next-7/13B, Qwen2-VL-2B/7B/72B. For LLaVA-1.5 series, we collect 
12,000 hallucinated caption instructions unlearning and another 36,000 visual reasoning samples from LLaVA-150K for the calculation of $\mathcal{L}_{\text{AR}}$. For the other models, the number of used unlearning and learning instructions is set to 2,000 and 8,000, respectively.

\paratitle{Benchmarks.} Aside from our constructed POPEv2, we also select other hallucination evaluation benchmarks for evaluation. We first select \textbf{POPE}~\cite{pope}, which detects hallucination by asking models whether the image contains a certain object. For generative evaluation benchmarks, we select \textbf{Object HalBench}~\cite{chair} and \textbf{MMHal-Bench}~\cite{llava_rlhf}. Object HalBench counts the hallucination objects and sentences in model-generated captions. MMHal-Bench adopts GPT-4 to evaluate fine-grained hallucination regarding counting and spatial relationships. We also include \textbf{LLaVA-Bench}~\cite{llava} for evaluating the open-ended generation abilities of models.

\paratitle{Baselines.} 
We compare our method with other hallucination mitigation methods. (1) Decoding-based methods, \eg VCD~\cite{vcd}, VDD~\cite{vdd} and OPERA~\cite{opera}. (2) Preference learning methods, \eg HA-DPO~\cite{ha-dpo} and POVID~\cite{povid}. These methods collect human preference data and adopt alignment algorithms like DPO~\cite{dpo} to align the model with human preference.

\subsection{Experiment Results}
We present the results of applying Obliviate to a diverse set of LVLMs in Table~\ref{tab:main}, compare it to other hallucination mitigation baselines in Table~\ref{tab:compare}, and evaluate its generalization capabilities in Table~\ref{tab:mme}.

\paratitle{Effectiveness across Various Model Families.}
On POPEv2, Obliviate consistently improves F1 Score and TNR, substantially reducing discriminative hallucinations. For example, LLaVA-1.5-7B’s TNR increases from 49.00\% to 84.20\%, indicating enhanced perception accuracy on counterfactual images and mitigation of training bias. The absolute PBO also decreases across all models, reflecting reduced response tendencies. On generative hallucination benchmarks like Object HalBench and MMHal-Bench, Obliviate effectively lowers hallucination rates while improving answer informativeness; for instance, LLaVA-Next-13B’s hallucination rate drops from 19.7 to 19.0, and its informativeness score rises from 2.91 to 3.10. Furthermore, on open-ended LLaVA-Bench, models with Obliviate consistently outperform vanilla models. Notably, Obliviate is effective across model sizes ranging from 2B to 72B. Qwen2-VL-7B with Obliviate even surpasses the larger Qwen2-VL-72B on most benchmarks.

\begin{table}[tbp]
    \centering
    \small
    \begin{tabular}{lrrcc}
    \toprule
    \multirow{2}{*}{\textbf{Method}} & \multirow{2}{*}{\textbf{\#Param}} & \multirow{2}{*}{\textbf{VRAM}} & \multirow{2}{*}{\shortstack{\textbf{GPU} \\ \textbf{Hours}}} & \multirow{2}{*}{\textbf{POPEv2}} \\
    & & & & \\
    \midrule
    LLaVA-NeXT$_\text{7B}$ & - & - & - &81.91    \\
    + SFT & 6.7 B & 77 GB & 26 &80.44    \\
    + LoRA & 0.02 B & 52 GB& 26 &77.59\\
    \rowcolor{gray!20}
    + Obliviate & 0.13 B & \textbf{28 GB} & \textbf{10} &\textbf{83.38}    \\
    \bottomrule
    \end{tabular}
    \caption{Comparison of computational efficiency between different training strategies on LLaVA-NeXT-7B.}
    \label{tab:efficiency}
\end{table}

\paratitle{Comparison with Baselines.} We compare Obliviate with other hallucination mitigation methods on LLaVA-1.5-7B and observe clear advantages. Unlike decoding-based approaches such as VCD and VDD, which impose constraints during inference and often yield limited gains or even performance drops on generation tasks, Obliviate consistently delivers stronger improvements. For example, it reduces the rate of hallucinated objects on Object HalBench from 25.1 to 18.3 and achieves a lower response error rate on MMHal-Bench (0.55), demonstrating its ability to mitigate hallucinations while preserving generation quality. Compared to preference learning methods like HA-DPO and POVID, Obliviate is also more cost-effective, as it does not require expensive human preference data or RL-style optimization, yet achieves comparable or even superior performance. On POPEv2, Obliviate achieves the highest F1 Score (82.91\%) and TNR (84.20\%) and matches the best LLaVA-Bench score (63.80), underscoring its effectiveness across different tasks.

\begin{table*}[tbp]
\centering
\scalebox{0.95}{
\begin{tabular}{lccccccccc}
\toprule 
\multirow{2.5}{*}{\textbf{Model}} & \multicolumn{3}{c}{\textbf{POPEv2}} &\textbf{POPE$_{adv}$} &\multicolumn{2}{c}{\textbf{Object HalBench}} &\multicolumn{2}{c}{\textbf{MMHal-Bench}} & \multirow{2.5}{*}{\textbf{LLaVA Bench}$\uparrow$ }\\
\cmidrule(lr){2-4}
\cmidrule(lr){5-5}
\cmidrule(lr){6-7} 
\cmidrule(lr){8-9} 
   & PBO & F1 Score $\uparrow$  & TNR $\uparrow$  & F1 Score $\uparrow$ & Resp. $\downarrow$ & Mention $\downarrow$  & Info. $\uparrow$  & Resp.$\downarrow$  \\
\midrule
LLaVA-1.5-7B & $+$ 23.20 & 77.44 & 49.00  & 77.57 &46.7&25.1& 2.19 &0.59 &61.50 \\
\midrule
+ VCD & $+$ 24.10 & 78.49 & 49.20 & 81.33 & 47.4 &25.2 & 2.12& 0.59&62.70\\
+ VDD & $+$ 20.30  &  78.30 & 50.80 & 82.22 & 46.7 &25.2 & \textbf{2.22} & \underline{0.56}& 63.20\\
+ OPERA &$+$ 19.10 & 79.28 & 58.60 & 81.88 & 45.1 &22.3 & 2.15 & 0.57 & 60.30\\
\midrule
+ HA-DPO &$+$ 20.40 & 78.41 & 53.60 & 82.05 & \underline{39.9} &\underline{19.9} & 1.98 & 0.60& \underline{63.50}\\
+ POVID & $+$ 16.80& \underline{80.53} & \underline{62.80}  & \underline{82.81} &48.1 &24.4 & 2.08 & \underline{0.56} & 62.20\\
\midrule
\rowcolor{gray!20}
+ Obliviate & $-$ 0.30 &\textbf{82.91} & \textbf{84.20}  & \textbf{84.50} &  \textbf{34.7} & \textbf{18.3} & \underline{2.20}& \textbf{0.55}& \textbf{63.80}  \\
\bottomrule
\end{tabular}}
\caption{Performance of representative hallucination mitigation methods and Obliviate.}
\label{tab:compare}
\end{table*}

\begin{table*}[tbp]
\centering
\small
\begin{tabular}{lccccccccc}
\toprule 
\multirow{2.5}{*}{\textbf{Model}} & \multicolumn{5}{c}{\textbf{MME$_{\text{perception}}$}} & \multicolumn{4}{c}{\textbf{ROPE}} \\
\cmidrule(lr){2-6}
\cmidrule(lr){7-10}
  & Count &  Position & Color & OCR & Sum & Wild  & Homo. & Adv. & Average  \\
\midrule
LLaVA-NeXT$_{\text{7B}}$ & 115 &125&115  &92.5& 1273.26 & 30.72 &66.69 & 45.34& 47.58\\
\rowcolor{gray!20}
+ Obliviate & \textbf{121.67} & \textbf{141.67}&\textbf{153.33}  &\textbf{152.5}&\textbf{1364.59} &\textbf{30.85} & \textbf{66.88} & \textbf{45.51} & \textbf{47.75} \\
\midrule
LLaVA-NeXT$_{\text{13B}}$ & 116.67  &115& \textbf{155}  &82.5& 1279.42 & 27.62	&63.23	&44.97	&45.27  \\
\rowcolor{gray!20}
+ Obliviate& \textbf{135}  &\textbf{126.67}&141.67  &\textbf{97.5}& \textbf{1352.64} & \textbf{28.29}	&\textbf{64.86}	&\textbf{45.03}	&\textbf{46.06} \\
\bottomrule
\end{tabular}
\caption{Evaluation results on the perception set of MME.}
\label{tab:mme}
\end{table*}

\begin{figure}[tbp]
    \centering
    \includegraphics[width=0.8\linewidth]{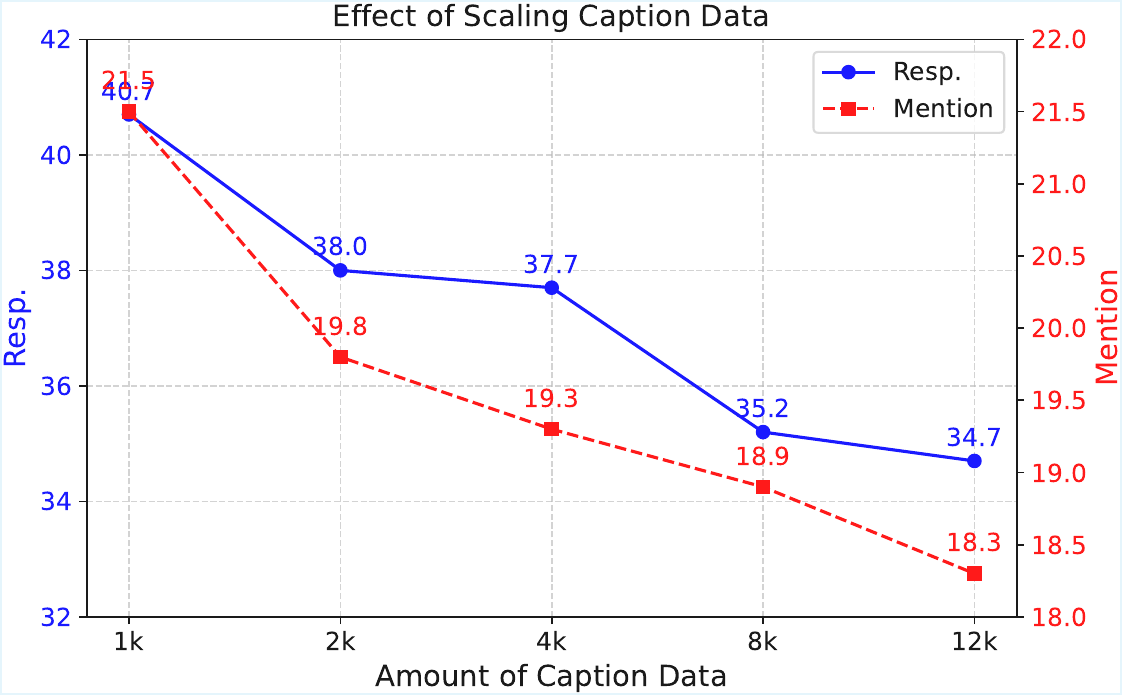}
    \caption{The effect of scaling caption data on the performance of LLaVA-1.5-7B on Object HalBench.}
    \label{fig:scaling}
\end{figure}

\paratitle{Generalization Capabilities.} We further evaluate the generalization ability of Obliviate on hallucination types beyond basic object hallucination. Specifically, we use the perception set of \textbf{MME}~\cite{fu2023mme}, which covers diverse perception challenges such as counting, positioning, color, and OCR, and the \textbf{ROPE}~\cite{chen2024rope} benchmark, which focuses on multi-object hallucination. As shown in Table~\ref{tab:mme}, Obliviate consistently improves baseline models across nearly all perception sub-tasks. For example, on LLaVA-NeXT-7B, it achieves significant gains in \emph{Position} (+16.67) and \emph{Color} (+38.33), raising the overall Sum score by 91.33 (1273.26 $\rightarrow$ 1364.59). Similar improvements are observed on Qwen2-VL-7B, where the Sum score increases from 1682.23 to 1691.68 with consistent gains across all categories. On the ROPE benchmark, Obliviate reduces hallucination across Wild, Homogeneous, and Adversarial settings for both 7B and 13B models. For instance, LLaVA-NeXT-13B improves its average score from 45.27 to 46.06, demonstrating stronger robustness against complex multi-object hallucinations. These results highlight Obliviate’s strong generalization to unseen hallucination types, suggesting that different hallucinations may stem from similar training biases.

\subsection{Further Analysis}

We analyze the computational efficiency and data scalability of Obliviate. We also present the ablation study on the unlearning factor and data ratio in the supplementary materials.

\paratitle{Computational Efficiency.}
Since Obliviate only updates the LM head of LVLMs, it offers clear advantages in computational efficiency. To demonstrate this, we compare full-parameter supervised fine-tuning (SFT), LoRA (rank = 8), and Obliviate on LLaVA-NeXT-7B. As shown in Table~\ref{tab:efficiency}. Obliviate achieves the best performance with the least memory and training time. Although LoRA has fewer trainable parameters, its VRAM consumption is nearly twice as high due to storing intermediate activations and gradients across layers. In contrast, Obliviate only computes gradients at the LM head, resulting in a simpler computation graph and much lower memory cost. Notably, both SFT and LoRA degrade model's performance, likely because unlearning across all layers disrupts the model’s generation patterns. In contrast, Obliviate targets only the LM head, effectively reducing hallucinations without harming generation quality.

\paratitle{Impact of Scaling Caption Data.}
We further investigate whether increasing the amount of caption data can enhance the performance of our method. To this end, we construct datasets containing 1k, 2k, 4k, and 8k hallucinated captions sampled from LLaVA-150K to fine-tune LLaVA-1.5. We then evaluate the resulting models on Object HalBench and present the results in Figure~\ref{fig:scaling}. The results indicate that as the volume of caption data increases, model hallucinations are further mitigated, demonstrating the strong scalability of our approach. We attribute this improvement to the fact that a larger number of hallucinated captions exposes the model to more biased generation patterns, enabling it to better unlearn these patterns and thereby reduce hallucinations.

\section{Related Work}

\paratitle{Large Vision-Language Models.}
LVLMs~\cite{llava15, minicpm-v, llava_ov} have demonstrated strong performance in multimodal tasks by integrating images into LLMs' embedding space using a visual encoder and projector. Recent advancements have enhanced their perceptual and cognitive abilities through various techniques. For instance, LLaVA-NeXT~\cite{llava_next} processes images at dynamic resolutions by dividing them into grids, while Qwen2-VL~\cite{qwen2vl} improves positional encoding for image and video inputs with M-RoPE. LVLMs can also handle both image and video modalities by unifying their processing and instruction mixing~\cite{llava_ov, internvl2}. Additionally, models like QvQ~\cite{qvq} and Virgo~\cite{virgo} have made significant progress in complex reasoning tasks by leveraging extended chain-of-thought reasoning.

\paratitle{Hallucination Mitigation for LVLMs.} Despite recent advancements, hallucination remains a significant challenge for LVLMs, impeding their accurate comprehension of visual inputs~\cite{wang2023evaluation, wang2023amber, chen2024rope}. Prior studies have explored various approaches to mitigate hallucinations in LVLMs~\cite{bpo, povid, ha-dpo, less_is_more, lancing, lrv, liu2024reducing, Xing2024efuf}. LLaVA-RLHF~\cite{llava_rlhf} and RLHF-V~\cite{rlhf-v} integrate preference learning techniques such as PPO~\cite{ppo} and DPO~\cite{dpo} into the LVLM training process, aligning model outputs with human to reduce hallucinations. In addition, several works have proposed training-free methods to enhance the decoding stage of LVLMs. VCD~\cite{vcd} and VDD~\cite{vdd} mitigate the influence of language priors by comparing the model’s original outputs to those generated with perturbed visual input. 
OPERA~\cite{opera} identifies a correlation between hallucinations and knowledge aggregation patterns in the self-attention matrix and introduces a rollback mechanism during decoding to prevent the generation of summary tokens.

\paratitle{Machine Unlearning.} Machine unlearning, initially developed for privacy protection, enables models to forget specific data without full retraining. It has been widely applied in various domains like image classification~\cite{Ginart-making-2019}, image generation~\cite{Gandikota-erasing-2023}, and LLMs~\cite{Liu-rethinking-2024}. Typical unlearning methods in LLMs involve directly training LLMs on specific dataset to erase knowledge, such as Gradient Ascent~\cite{Yao-unlearning-2024} and NPO~\cite{npo}. By designing special training objectives (\eg loss maximization), unlearning methods are more efficient solutions to reduce the memorization of specific data~\cite{npo,Liu-rethinking-2024}.
In this work, we apply machine unlearning techniques on LVLMs for hallucination mitigation by employing gradient ascent.

\section{Conclusion}
In this work, we systematically examined the role of training data in hallucination for LVLMs by the proposed benchmark POPEv2 and revealed that LVLMs exhibit a clear training bias: they hallucinate more frequently even on images seen during training. Through probing experiments, we identified that this bias is mainly rooted in the LM head, which fails to translate accurate visual representations into correct textual outputs. To mitigate this issue, we introduced Obliviate, a lightweight and parameter-efficient unlearning method that targets training bias by only fine-tuning the LM head with biased inference results derived from the original training data. Extensive experiments demonstrated that Obliviate significantly reduces hallucination across different tasks, while maintaining scalability across model sizes and data volumes. Beyond object-level hallucination, our method also shows promising generalization to other hallucination types, providing a new perspective on leveraging training data to enhance the reliability of LVLMs.

\newpage
\section{More Details on Data Collection Process}
To avoid trivial masks on unimportant objects or aggressive masks that remove too much visual information, we constrain the size of the target object to 5\%–25\% of the image size. If multiple objects satisfy the criteria, we select the least frequent one in the collected data to ensure a balanced distribution of object classes. Once the target object is identified, it is replaced by a black mask using the bounding box provided in the MSCOCO annotations. Finally, we employ human annotators to verify the quality of counterfactual images, ensuring that the target object is no longer visible.

\begin{figure}[tbp]
    \centering
    \begin{subfigure}{0.49\linewidth}
        \centering
        \includegraphics[width=\linewidth]{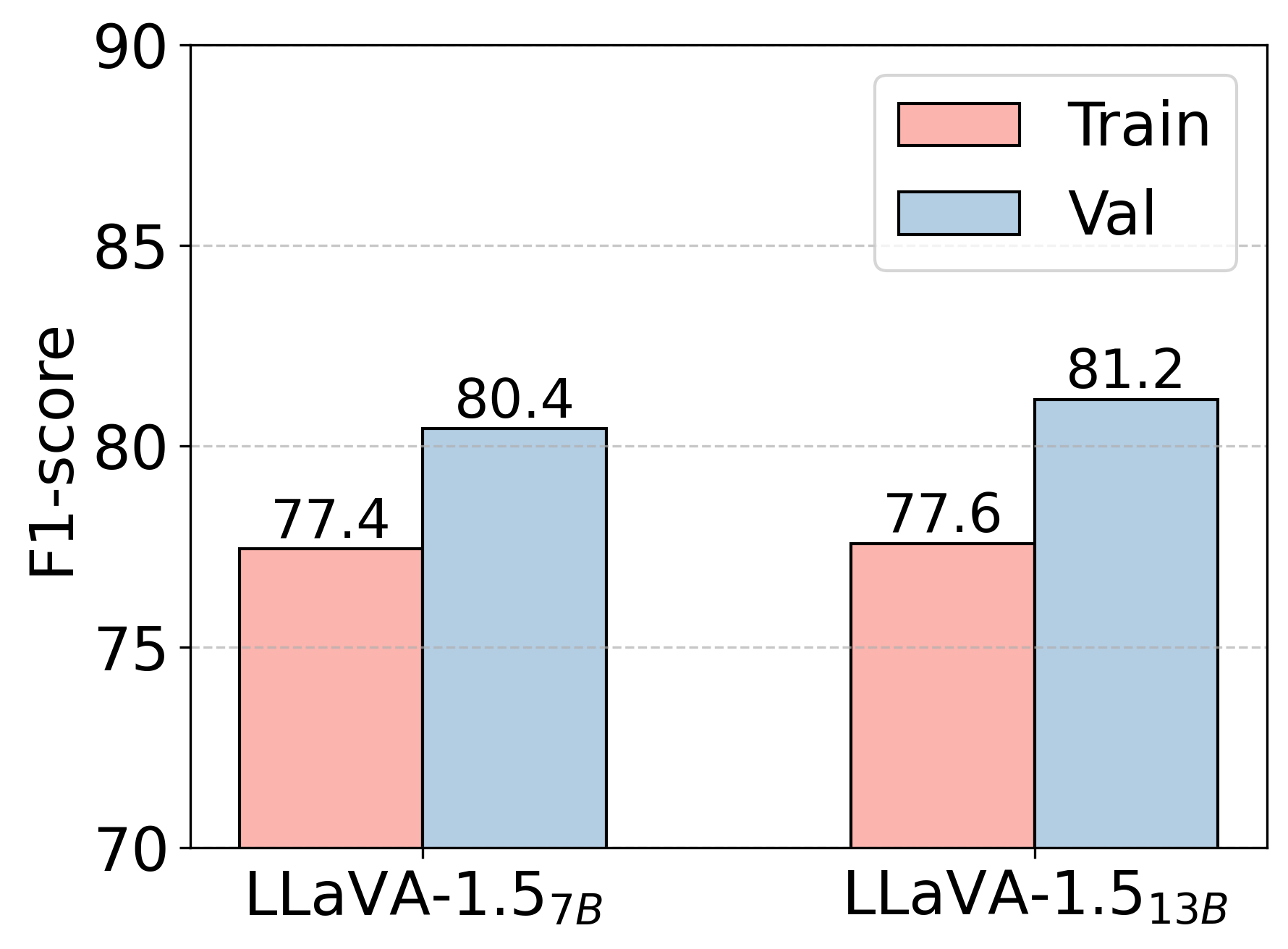}
        \caption{Train \emph{vs.} Validation} \label{fig:comparison1}
        \label{fig:sub1}
    \end{subfigure}
    \begin{subfigure}{0.49\linewidth}
        \centering
        \includegraphics[width=\linewidth]{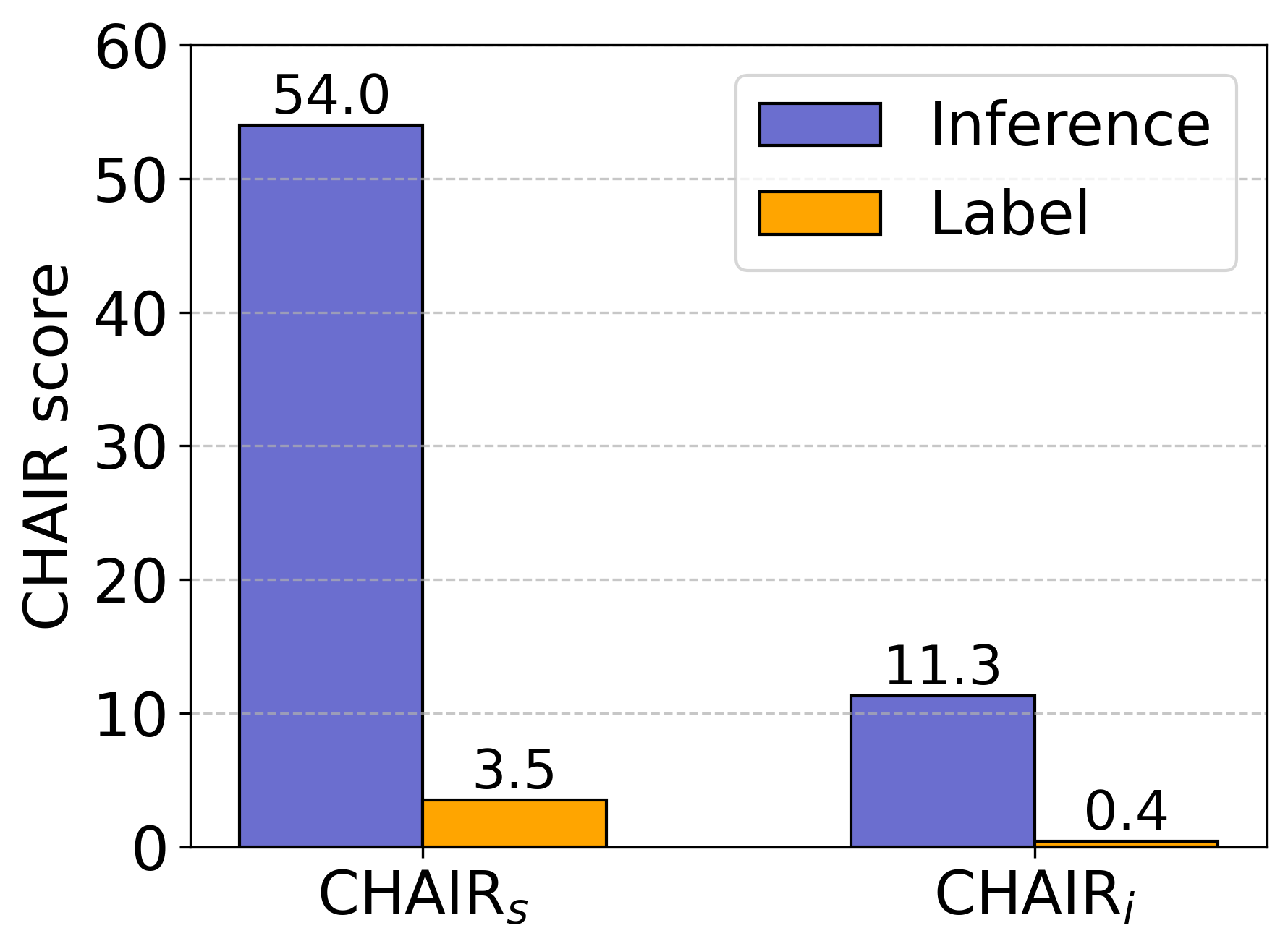}
        \caption{Inference \emph{vs.} Label} \label{fig:comparison2}
        \label{fig:sub2}
    \end{subfigure}
    \caption{Hallucination analysis in LLaVA-1.5: (a) Performance on seen vs. unseen images, and (b) Regenerated captions vs. original training labels.}
\end{figure}

\section{Additional Experiments on the Influence of Training Data on Hallucination}
We conduct two additional experiments to further examine how training data influences hallucination. First, using the same pipeline as POPEv2, we collect new data from the MSCOCO 2017 validation set and evaluate LLaVA-1.5. As shown in Figure~\ref{fig:comparison1}, the model achieves a higher F1-score on these previously unseen images. Next, we filter caption instructions from LLaVA-1.5’s training data and let the model to regenerate captions on these data. Using the CHAIR metric~\cite{chair}, we compare hallucination levels in ground-truth labels and inference results. As illustrated in  Figure~\ref{fig:comparison2}, while the training labels are nearly hallucination-free, LLaVA-1.5 produces captions with significant hallucinations, even has encountered these images during training. These findings suggest that current LVLMs commonly suffer from training bias, which may stem from the overfitting to specific patterns in the training data. This bias could lead the model to rely excessively on learned priors rather than accurately interpreting the visual input, thereby increasing hallucination rates even on familiar data.

\begin{table}[tbp]
    \centering
    \begin{tabular}{cccccc}
    \toprule
    \multirow{2.5}{*}{\textbf{Config}} & \multicolumn{2}{c}{\textbf{POPEv2}} & \multicolumn{2}{c}{\textbf{Object HalBench}}  \\
    \cmidrule(lr){2-3} 
    \cmidrule(lr){4-5}
    & F1 Score  & TNR  & Resp.  & Mention \\
    \midrule
    2k-8k, $\alpha=0.01$ & 84.54 & 70.80    & 43.0 & 22.5 \\
    2k-8k, $\alpha=0.02$ & 87.38  & 82.20   & 38.0 & 19.8 \\
    2k-8k, $\alpha=0.05$ & 83.57 &  66.80  & 49.8  & 24.9 \\
    \bottomrule
    \end{tabular}
    \caption{Impact of  different unlearning factor $\alpha$.}
    \label{tab:alpha}
\end{table}
\section{Implementation Details for Probing Experiment}
For the data preparation., we select ``\emph{dog}'' as the target object for probing and take MSCOCO 2017 as our image source. We first collect 7,000 images with half of the images containing ``\emph{dog}'' and the other half not. We then split images into train and validation sets at a ratio of 7:1. For the test set, our goal is to evaluate whether the probes trained on standard images can accurately determine the absence of an object in counterfactual images. Therefore, we construct another 177 counterfactual images by masking the dog contained in each of them. We select LLaVA-1.5-7B as the target model and set the $I$ as the images from our training set and $T$ as ``\emph{Is there a dog in the image?}''. We then extract the encoded image features \( E_I \), the image embeddings \( H_I^0 \), and the image and text hidden states from layers 1, 16, and 32 of the model. These features are used to train the linear probe, and we select the checkpoint according to the performance on validation set.

\section{Ablation on Unlearning Factor $\alpha$}
We also analyze the effect of the unlearning parameter $\alpha$  (set to 0.01, 0.02, and 0.05) on the performance of LLaVA-1.5-7B. Results in Table~\ref{tab:alpha} show that even a small $\alpha$  (0.01) effectively reduces hallucinations. Increasing $\alpha$  to 0.02 further mitigates them, but at $\alpha$  = 0.05, performance degrades, with occasional repetitive outputs. A possible reason is that the excessive unlearning process would disrupt the model’s original generative capabilities, underscoring the need for careful $\alpha$ calibration.

\section{Ablation on Data Ratio}
Finally, we investigate the impact of the ratio between caption data and general visual instruction data (for gradient descent only) on the performance of LLaVA-1.5-7B. We evaluate three distinct data configurations (\ie, 2k-2k, 2k-8k, and 2k-16k mixtures of caption and general instruction data) while training the models for the same number of steps. As shown in Table~\ref{tab:ratio}, Obliviate demonstrates strong robustness, effectively mitigating hallucinations across all data ratios. Furthermore, we observe that as the proportion of caption data used for unlearning increases, Obliviate becomes increasingly effective in reducing generative hallucinations. This finding further validates the efficacy of the bias unlearning strategy, which can work well using less data.

\begin{table}[tbp]
    \centering
    \begin{tabular}{cccccc}
    \toprule
    \multirow{2.5}{*}{\textbf{Config}} & \multicolumn{2}{c}{\textbf{POPEv2}} & \multicolumn{2}{c}{\textbf{Object HalBench}}  \\
    \cmidrule(lr){2-3} 
    \cmidrule(lr){4-5}
    & F1-score  & TNR  & Resp.  & Mention \\
    \midrule
    2k-2k & 86.49  & 87.40   &36.3 & 20.3 \\
    2k-8k & 87.38  & 82.20   & 38.0 & 19.8 \\
    2k-16k & 85.83  & 88.00   & 42.7 & 21.8 \\
    \bottomrule
    \end{tabular}
    \caption{Impact of different data ratios. }
    \label{tab:ratio}
\end{table}


\begin{thebibliography}{42}
\providecommand{\natexlab}[1]{#1}

\bibitem[{Alain and Bengio(2017)}]{probe}
Alain, G.; and Bengio, Y. 2017.
\newblock Understanding intermediate layers using linear classifier probes.
\newblock In \emph{{ICLR} (Workshop)}. OpenReview.net.

\bibitem[{Chen et~al.(2024{\natexlab{a}})Chen, Li, Dong, Zhang, He, Wang, Zhao, and Lin}]{sharegpt4v}
Chen, L.; Li, J.; Dong, X.; Zhang, P.; He, C.; Wang, J.; Zhao, F.; and Lin, D. 2024{\natexlab{a}}.
\newblock ShareGPT4V: Improving Large Multi-modal Models with Better Captions.
\newblock In \emph{{ECCV} {(17)}}, volume 15075 of \emph{Lecture Notes in Computer Science}, 370--387. Springer.

\bibitem[{Chen et~al.(2024{\natexlab{b}})Chen, Ma, Zhang, Xu, Qian, Yang, Fouhey, and Chai}]{chen2024rope}
Chen, X.; Ma, Z.; Zhang, X.; Xu, S.; Qian, S.; Yang, J.; Fouhey, D.; and Chai, J. 2024{\natexlab{b}}.
\newblock Multi-Object Hallucination in Vision Language Models.
\newblock In \emph{NeurIPS}.

\bibitem[{Chen et~al.(2024{\natexlab{c}})Chen, Wang, Cao, Liu, Gao, Cui, Zhu, Ye, Tian, Liu, Gu, Wang, Li, Ren, Chen, Luo, Wang, Jiang, Wang, He, Shi, Zhang, Lv, Wang, Shao, Chu, Tu, He, Wu, Deng, Ge, Chen, Dou, Lu, Zhu, Lu, Lin, Qiao, Dai, and Wang}]{internvl-2.5}
Chen, Z.; Wang, W.; Cao, Y.; Liu, Y.; Gao, Z.; Cui, E.; Zhu, J.; Ye, S.; Tian, H.; Liu, Z.; Gu, L.; Wang, X.; Li, Q.; Ren, Y.; Chen, Z.; Luo, J.; Wang, J.; Jiang, T.; Wang, B.; He, C.; Shi, B.; Zhang, X.; Lv, H.; Wang, Y.; Shao, W.; Chu, P.; Tu, Z.; He, T.; Wu, Z.; Deng, H.; Ge, J.; Chen, K.; Dou, M.; Lu, L.; Zhu, X.; Lu, T.; Lin, D.; Qiao, Y.; Dai, J.; and Wang, W. 2024{\natexlab{c}}.
\newblock Expanding Performance Boundaries of Open-Source Multimodal Models with Model, Data, and Test-Time Scaling.
\newblock \emph{CoRR}, abs/2412.05271.

\bibitem[{Chen et~al.(2024{\natexlab{d}})Chen, Wang, Tian, Ye, Gao, Cui, Tong, Hu, Luo, Ma, Ma, Wang, Dong, Yan, Guo, He, Shi, Jin, Xu, Wang, Wei, Li, Zhang, Zhang, Cai, Wen, Yan, Dou, Lu, Zhu, Lu, Lin, Qiao, Dai, and Wang}]{internvl2}
Chen, Z.; Wang, W.; Tian, H.; Ye, S.; Gao, Z.; Cui, E.; Tong, W.; Hu, K.; Luo, J.; Ma, Z.; Ma, J.; Wang, J.; Dong, X.; Yan, H.; Guo, H.; He, C.; Shi, B.; Jin, Z.; Xu, C.; Wang, B.; Wei, X.; Li, W.; Zhang, W.; Zhang, B.; Cai, P.; Wen, L.; Yan, X.; Dou, M.; Lu, L.; Zhu, X.; Lu, T.; Lin, D.; Qiao, Y.; Dai, J.; and Wang, W. 2024{\natexlab{d}}.
\newblock How Far Are We to GPT-4V? Closing the Gap to Commercial Multimodal Models with Open-Source Suites.
\newblock \emph{CoRR}, abs/2404.16821.

\bibitem[{Du et~al.(2025)Du, Liu, Li, Zhao, Huo, Wang, Chen, Liu, Wang, and Wen}]{virgo}
Du, Y.; Liu, Z.; Li, Y.; Zhao, W.~X.; Huo, Y.; Wang, B.; Chen, W.; Liu, Z.; Wang, Z.; and Wen, J.-R. 2025.
\newblock Virgo: A Preliminary Exploration on Reproducing o1-like MLLM.
\newblock \emph{CoRR}, abs/2501.01904.

\bibitem[{Fu et~al.(2023)Fu, Chen, Shen, Qin, Zhang, Lin, Qiu, Lin, Yang, Zheng, Li, Sun, and Ji}]{fu2023mme}
Fu, C.; Chen, P.; Shen, Y.; Qin, Y.; Zhang, M.; Lin, X.; Qiu, Z.; Lin, W.; Yang, J.; Zheng, X.; Li, K.; Sun, X.; and Ji, R. 2023.
\newblock {MME:} {A} Comprehensive Evaluation Benchmark for Multimodal Large Language Models.
\newblock \emph{CoRR}, abs/2306.13394.

\bibitem[{Gandikota et~al.(2023)Gandikota, Materzynska, Fiotto{-}Kaufman, and Bau}]{Gandikota-erasing-2023}
Gandikota, R.; Materzynska, J.; Fiotto{-}Kaufman, J.; and Bau, D. 2023.
\newblock Erasing Concepts from Diffusion Models.
\newblock In \emph{{ICCV}}, 2426--2436. {IEEE}.

\bibitem[{Ginart et~al.(2019)Ginart, Guan, Valiant, and Zou}]{Ginart-making-2019}
Ginart, A.; Guan, M.~Y.; Valiant, G.; and Zou, J. 2019.
\newblock Making {AI} Forget You: Data Deletion in Machine Learning.
\newblock In \emph{NeurIPS}, 3513--3526.

\bibitem[{Huang et~al.(2024)Huang, Dong, Zhang, Wang, He, Wang, Lin, Zhang, and Yu}]{opera}
Huang, Q.; Dong, X.; Zhang, P.; Wang, B.; He, C.; Wang, J.; Lin, D.; Zhang, W.; and Yu, N. 2024.
\newblock {OPERA:} Alleviating Hallucination in Multi-Modal Large Language Models via Over-Trust Penalty and Retrospection-Allocation.
\newblock In \emph{{CVPR}}, 13418--13427. {IEEE}.

\bibitem[{Leng et~al.(2024)Leng, Zhang, Chen, Li, Lu, Miao, and Bing}]{vcd}
Leng, S.; Zhang, H.; Chen, G.; Li, X.; Lu, S.; Miao, C.; and Bing, L. 2024.
\newblock Mitigating Object Hallucinations in Large Vision-Language Models through Visual Contrastive Decoding.
\newblock In \emph{{CVPR}}, 13872--13882. {IEEE}.

\bibitem[{Li et~al.(2024{\natexlab{a}})Li, Zhang, Guo, Zhang, Li, Zhang, Zhang, Li, Liu, and Li}]{llava_ov}
Li, B.; Zhang, Y.; Guo, D.; Zhang, R.; Li, F.; Zhang, H.; Zhang, K.; Li, Y.; Liu, Z.; and Li, C. 2024{\natexlab{a}}.
\newblock LLaVA-OneVision: Easy Visual Task Transfer.
\newblock \emph{CoRR}, abs/2408.03326.

\bibitem[{Li et~al.(2024{\natexlab{b}})Li, Zhang, Geng, Geng, Long, Shen, Zhang, Liu, and Dong}]{manipllm}
Li, X.; Zhang, M.; Geng, Y.; Geng, H.; Long, Y.; Shen, Y.; Zhang, R.; Liu, J.; and Dong, H. 2024{\natexlab{b}}.
\newblock ManipLLM: Embodied Multimodal Large Language Model for Object-Centric Robotic Manipulation.
\newblock In \emph{{CVPR}}, 18061--18070. {IEEE}.

\bibitem[{Li et~al.(2023)Li, Du, Zhou, Wang, Zhao, and Wen}]{pope}
Li, Y.; Du, Y.; Zhou, K.; Wang, J.; Zhao, W.~X.; and Wen, J. 2023.
\newblock Evaluating Object Hallucination in Large Vision-Language Models.
\newblock In \emph{{EMNLP}}, 292--305. Association for Computational Linguistics.

\bibitem[{Lin et~al.(2014)Lin, Maire, Belongie, Hays, Perona, Ramanan, Doll{\'{a}}r, and Zitnick}]{coco}
Lin, T.; Maire, M.; Belongie, S.~J.; Hays, J.; Perona, P.; Ramanan, D.; Doll{\'{a}}r, P.; and Zitnick, C.~L. 2014.
\newblock Microsoft {COCO:} Common Objects in Context.
\newblock In \emph{{ECCV} {(5)}}, volume 8693 of \emph{Lecture Notes in Computer Science}, 740--755. Springer.

\bibitem[{Liu et~al.(2024{\natexlab{a}})Liu, Lin, Li, Wang, Yacoob, and Wang}]{lrv}
Liu, F.; Lin, K.; Li, L.; Wang, J.; Yacoob, Y.; and Wang, L. 2024{\natexlab{a}}.
\newblock Mitigating Hallucination in Large Multi-Modal Models via Robust Instruction Tuning.
\newblock In \emph{{ICLR}}. OpenReview.net.

\bibitem[{Liu et~al.(2024{\natexlab{b}})Liu, Li, Li, and Lee}]{llava15}
Liu, H.; Li, C.; Li, Y.; and Lee, Y.~J. 2024{\natexlab{b}}.
\newblock Improved Baselines with Visual Instruction Tuning.
\newblock In \emph{{CVPR}}, 26286--26296. {IEEE}.

\bibitem[{Liu et~al.(2024{\natexlab{c}})Liu, Li, Li, Li, Zhang, Shen, and Lee}]{llava_next}
Liu, H.; Li, C.; Li, Y.; Li, B.; Zhang, Y.; Shen, S.; and Lee, Y.~J. 2024{\natexlab{c}}.
\newblock LLaVA-NeXT: Improved reasoning, OCR, and world knowledge.

\bibitem[{Liu et~al.(2023)Liu, Li, Wu, and Lee}]{llava}
Liu, H.; Li, C.; Wu, Q.; and Lee, Y.~J. 2023.
\newblock Visual Instruction Tuning.
\newblock In \emph{NeurIPS}.

\bibitem[{Liu et~al.(2024{\natexlab{d}})Liu, Yao, Jia, Casper, Baracaldo, Hase, Xu, Yao, Li, Varshney, Bansal, Koyejo, and Liu}]{Liu-rethinking-2024}
Liu, S.; Yao, Y.; Jia, J.; Casper, S.; Baracaldo, N.; Hase, P.; Xu, X.; Yao, Y.; Li, H.; Varshney, K.~R.; Bansal, M.; Koyejo, S.; and Liu, Y. 2024{\natexlab{d}}.
\newblock Rethinking Machine Unlearning for Large Language Models.
\newblock \emph{CoRR}, abs/2402.08787.

\bibitem[{Liu et~al.(2024{\natexlab{e}})Liu, Ye, Xing, and Zou}]{liu2024reducing}
Liu, S.; Ye, H.; Xing, L.; and Zou, J. 2024{\natexlab{e}}.
\newblock Reducing hallucinations in vision-language models via latent space steering.
\newblock \emph{arXiv preprint arXiv:2410.15778}.

\bibitem[{OpenAI(2023)}]{gpt-4v}
OpenAI. 2023.
\newblock GPT-4V(ision) System Card.

\bibitem[{Pi et~al.(2024)Pi, Han, Xiong, Zhang, Liu, Pan, and Zhang}]{bpo}
Pi, R.; Han, T.; Xiong, W.; Zhang, J.; Liu, R.; Pan, R.; and Zhang, T. 2024.
\newblock Strengthening multimodal large language model with bootstrapped preference optimization.
\newblock In \emph{European Conference on Computer Vision}, 382--398. Springer.

\bibitem[{{Qwen Team}(2024)}]{qvq}
{Qwen Team}. 2024.
\newblock QVQ: To See the World with Wisdom.

\bibitem[{Rafailov et~al.(2023)Rafailov, Sharma, Mitchell, Manning, Ermon, and Finn}]{dpo}
Rafailov, R.; Sharma, A.; Mitchell, E.; Manning, C.~D.; Ermon, S.; and Finn, C. 2023.
\newblock Direct Preference Optimization: Your Language Model is Secretly a Reward Model.
\newblock In \emph{NeurIPS}.

\bibitem[{Rohrbach et~al.(2018)Rohrbach, Hendricks, Burns, Darrell, and Saenko}]{chair}
Rohrbach, A.; Hendricks, L.~A.; Burns, K.; Darrell, T.; and Saenko, K. 2018.
\newblock Object Hallucination in Image Captioning.
\newblock In \emph{{EMNLP}}, 4035--4045. Association for Computational Linguistics.

\bibitem[{Schulman et~al.(2017)Schulman, Wolski, Dhariwal, Radford, and Klimov}]{ppo}
Schulman, J.; Wolski, F.; Dhariwal, P.; Radford, A.; and Klimov, O. 2017.
\newblock Proximal Policy Optimization Algorithms.
\newblock \emph{CoRR}, abs/1707.06347.

\bibitem[{Sun et~al.(2024)Sun, Shen, Cao, Liu, Li, Shen, Gan, Gui, Wang, Yang, Keutzer, and Darrell}]{llava_rlhf}
Sun, Z.; Shen, S.; Cao, S.; Liu, H.; Li, C.; Shen, Y.; Gan, C.; Gui, L.; Wang, Y.; Yang, Y.; Keutzer, K.; and Darrell, T. 2024.
\newblock Aligning Large Multimodal Models with Factually Augmented {RLHF}.
\newblock In \emph{{ACL} (Findings)}, 13088--13110. Association for Computational Linguistics.

\bibitem[{Tong et~al.(2024)Tong, Liu, Zhai, Ma, LeCun, and Xie}]{mmvp}
Tong, S.; Liu, Z.; Zhai, Y.; Ma, Y.; LeCun, Y.; and Xie, S. 2024.
\newblock Eyes Wide Shut? Exploring the Visual Shortcomings of Multimodal LLMs.
\newblock In \emph{{CVPR}}, 9568--9578. {IEEE}.

\bibitem[{Wang et~al.(2023{\natexlab{a}})Wang, Wang, Xu, Zhang, Gu, Jia, Wang, Xu, Yan, Zhang et~al.}]{wang2023amber}
Wang, J.; Wang, Y.; Xu, G.; Zhang, J.; Gu, Y.; Jia, H.; Wang, J.; Xu, H.; Yan, M.; Zhang, J.; et~al. 2023{\natexlab{a}}.
\newblock Amber: An llm-free multi-dimensional benchmark for mllms hallucination evaluation.
\newblock \emph{arXiv preprint arXiv:2311.07397}.

\bibitem[{Wang et~al.(2023{\natexlab{b}})Wang, Zhou, Xu, Shi, Zhao, Xu, Ye, Yan, Zhang, Zhu et~al.}]{wang2023evaluation}
Wang, J.; Zhou, Y.; Xu, G.; Shi, P.; Zhao, C.; Xu, H.; Ye, Q.; Yan, M.; Zhang, J.; Zhu, J.; et~al. 2023{\natexlab{b}}.
\newblock Evaluation and analysis of hallucination in large vision-language models.
\newblock \emph{arXiv preprint arXiv:2308.15126}.

\bibitem[{Wang et~al.(2024)Wang, Bai, Tan, Wang, Fan, Bai, Chen, Liu, Wang, Ge, Fan, Dang, Du, Ren, Men, Liu, Zhou, Zhou, and Lin}]{qwen2vl}
Wang, P.; Bai, S.; Tan, S.; Wang, S.; Fan, Z.; Bai, J.; Chen, K.; Liu, X.; Wang, J.; Ge, W.; Fan, Y.; Dang, K.; Du, M.; Ren, X.; Men, R.; Liu, D.; Zhou, C.; Zhou, J.; and Lin, J. 2024.
\newblock Qwen2-VL: Enhancing Vision-Language Model's Perception of the World at Any Resolution.
\newblock \emph{CoRR}, abs/2409.12191.

\bibitem[{Xing et~al.(2024)Xing, Zhao, Wu, An, Chen, Li, Zhang, and Dai}]{Xing2024efuf}
Xing, S.; Zhao, F.; Wu, Z.; An, T.; Chen, W.; Li, C.; Zhang, J.; and Dai, X. 2024.
\newblock {EFUF:} Efficient Fine-Grained Unlearning Framework for Mitigating Hallucinations in Multimodal Large Language Models.
\newblock In \emph{{EMNLP}}, 1167--1181. Association for Computational Linguistics.

\bibitem[{Yao, Xu, and Liu(2024)}]{Yao-unlearning-2024}
Yao, Y.; Xu, X.; and Liu, Y. 2024.
\newblock Large Language Model Unlearning.
\newblock In \emph{NeurIPS}.

\bibitem[{Yao et~al.(2024)Yao, Yu, Zhang, Wang, Cui, Zhu, Cai, Li, Zhao, He, Chen, Zhou, Zou, Zhang, Hu, Zheng, Zhou, Cai, Han, Zeng, Li, Liu, and Sun}]{minicpm-v}
Yao, Y.; Yu, T.; Zhang, A.; Wang, C.; Cui, J.; Zhu, H.; Cai, T.; Li, H.; Zhao, W.; He, Z.; Chen, Q.; Zhou, H.; Zou, Z.; Zhang, H.; Hu, S.; Zheng, Z.; Zhou, J.; Cai, J.; Han, X.; Zeng, G.; Li, D.; Liu, Z.; and Sun, M. 2024.
\newblock MiniCPM-V: {A} {GPT-4V} Level {MLLM} on Your Phone.
\newblock \emph{CoRR}, abs/2408.01800.

\bibitem[{Yu et~al.(2024)Yu, Yao, Zhang, He, Han, Cui, Hu, Liu, Zheng, and Sun}]{rlhf-v}
Yu, T.; Yao, Y.; Zhang, H.; He, T.; Han, Y.; Cui, G.; Hu, J.; Liu, Z.; Zheng, H.; and Sun, M. 2024.
\newblock {RLHF-V:} Towards Trustworthy MLLMs via Behavior Alignment from Fine-Grained Correctional Human Feedback.
\newblock In \emph{{CVPR}}, 13807--13816. {IEEE}.

\bibitem[{Yue, Zhang, and Jin(2024)}]{less_is_more}
Yue, Z.; Zhang, L.; and Jin, Q. 2024.
\newblock Less is More: Mitigating Multimodal Hallucination from an {EOS} Decision Perspective.
\newblock In \emph{{ACL} {(1)}}, 11766--11781. Association for Computational Linguistics.

\bibitem[{Zhang et~al.(2024{\natexlab{a}})Zhang, Lin, Bai, and Mei}]{npo}
Zhang, R.; Lin, L.; Bai, Y.; and Mei, S. 2024{\natexlab{a}}.
\newblock Negative Preference Optimization: From Catastrophic Collapse to Effective Unlearning.
\newblock \emph{CoRR}, abs/2404.05868.

\bibitem[{Zhang et~al.(2024{\natexlab{b}})Zhang, Yu, Wen, Wang, Zhang, Wang, Jin, and Tan}]{vdd}
Zhang, Y.; Yu, W.; Wen, Q.; Wang, X.; Zhang, Z.; Wang, L.; Jin, R.; and Tan, T. 2024{\natexlab{b}}.
\newblock Debiasing Multimodal Large Language Models.
\newblock \emph{CoRR}, abs/2403.05262.

\bibitem[{Zhao et~al.(2024)Zhao, Si, Chen, Zhang, Sun, Zhang, and Chang}]{lancing}
Zhao, H.; Si, S.; Chen, L.; Zhang, Y.; Sun, M.; Zhang, M.; and Chang, B. 2024.
\newblock Looking Beyond Text: Reducing Language bias in Large Vision-Language Models via Multimodal Dual-Attention and Soft-Image Guidance.
\newblock \emph{CoRR}, abs/2411.14279.

\bibitem[{Zhao et~al.(2023)Zhao, Wang, Ouyang, Dong, Wang, and He}]{ha-dpo}
Zhao, Z.; Wang, B.; Ouyang, L.; Dong, X.; Wang, J.; and He, C. 2023.
\newblock Beyond Hallucinations: Enhancing LVLMs through Hallucination-Aware Direct Preference Optimization.
\newblock \emph{CoRR}, abs/2311.16839.

\bibitem[{Zhou et~al.(2024)Zhou, Cui, Rafailov, Finn, and Yao}]{povid}
Zhou, Y.; Cui, C.; Rafailov, R.; Finn, C.; and Yao, H. 2024.
\newblock Aligning Modalities in Vision Large Language Models via Preference Fine-tuning.
\newblock \emph{CoRR}, abs/2402.11411.

\end{thebibliography}
\end{document}